\documentclass[10pt,twocolumn,letterpaper]{article}

\usepackage{booktabs}
\usepackage{times}
\usepackage{epsfig}
\usepackage{graphicx}
\usepackage{amsmath}
\usepackage{amssymb}
\usepackage{amsthm}
\usepackage{array}
\newcolumntype{P}[1]{>{\centering\arraybackslash}p{#1}}

\usepackage[final]{cvpr}
\makeatletter
\@namedef{ver@everyshi.sty}{}
\makeatother      %
\usepackage[pagebackref,breaklinks,colorlinks]{hyperref}
\usepackage[capitalize]{cleveref}
\crefname{section}{Sec.}{Secs.}
\Crefname{section}{Section}{Sections}
\Crefname{table}{Table}{Tables}
\crefname{table}{Tab.}{Tabs.}
\def\cvprPaperID{2898} %
\def\confName{CVPR}
\def\confYear{2022}

\usepackage[inkscapelatex=false]{svg}

\newcommand{\jtr}{character-context decoupling}
\newcommand{\Jtr}{Character-Context Decoupling}

\newcommand{\jtrf}{character-context decoupling framework}
\newcommand{\JtrF}{Character-Context Decoupling Framework}

\newcommand{\Jtrn}{Open-set Character-Context Decoupling network}
\newcommand{\JTRN}{OpenCCD}

\newcommand{\gta}{detached temporal attention}
\newcommand{\Gta}{Detached Temporal Attention}

\newcommand{\Ca}{Decoupled Context Anchor}

\newcommand{\tvi}{{character visual information}}
\newcommand{\TvI}{{Character Visual Information}}
\newcommand{\tsi}{{contextual information}}
\newcommand{\Tsi}{{Contextual information}}
\newcommand{\tti}{{temporal information}}

\newcommand{\Tli}{{Linguistic information}}
\newcommand{\tli}{{linguistic information}}
\newcommand{\TlI}{{Linguistic Information}}

\newcommand{\idx}[2]{#1_{[#2]}}
\newcommand{\predseq}{\boldsymbol{\hat{y}}}
\newcommand{\anyseq}{\boldsymbol{y}}
\newcommand{\predat}[1]{\idx{\hat{y}}{#1}}

\newcommand{\charset}{\mathcal{Y}}
\newcommand{\predalterat}[1]{\idx{y^{'}}{#1}}

\newcommand{\predchanya}{\anypredat{t}}

\newcommand{\anypredseq}{\boldsymbol{y}}
\newcommand{\anypredat}[1]{\idx{y}{#1}}
\newcommand{\anydistat}[1]{\idx{\mathbf{Y}}{#1}}

\newcommand{\gtseq}{\boldsymbol{y^{*}}}
\newcommand{\gtat}[1]{\idx{y^{*}}{#1}}

\newcommand{\imgword}{\boldsymbol{x}}
\newcommand{\imgat}[1]{\idx{x}{#1}}
\newcommand{\imgchany}{\imgat{t}}
\newcommand{\imgchanya}{\imgchany}

\newcommand{\ctxat}[1]{c}
\newcommand{\ctxestat}[1]{\hat{c}}
\newcommand{\ctxdist}[1]{\idx{C}{#1}}

\newcommand{\ctxchanya}{c}

\newcommand{\anchoreddist}[1]{\int_{}^{\ctxat{t} \in \ctxdist{t}}{#1}}
\newcommand{\templates}{E}
\newcommand{\dcarightfull}{\beta(\textbf{y})\prod_{t=1}^lP(\anypredat{t}|\imgat{t})\prod_{t=1}^l\anchoreddist{P(\predchanya|\ctxchanya) P(\ctxat{t}|\imgword,l)}}
\newcommand{\dcaright}{\prod_{t=1}^lP(\anypredat{t}|\imgat{t})\prod_{t=1}^l\anchoreddist{P(\predchanya|\ctxchanya) P(\ctxat{t}|\imgword,l)}}

\newcounter{relctr} %
\everydisplay\expandafter{\the\everydisplay\setcounter{relctr}{0}} %

\newcommand\labelrel[2]{%
	\begingroup
	\refstepcounter{relctr}%
	\stackrel{\textnormal{(\alph{relctr})}}{\mathstrut{#1}}%
	\originallabel{#2}%
	\endgroup
}

\newcommand{\chgninC}[1]{{\color{black}{#1}}}
\newcommand{\chgeigC}[1]{{\color{black}{#1}}}
\newcommand{\chgfiveC}[1]{{\color{black}{#1}}}
\newcommand{\chgsevC}[1]{{\color{black}{#1}}}

\newcommand{\chgrevA}[1]{{\color{black}{#1}}}
\newcommand{\chgrevC}[1]{{\color{black}{#1}}}
\newcommand{\chgrevD}[1]{{\color{black}{#1}}}
\newcommand{\chgclarify}[1]{{\color{black}{#1}}}

\AtBeginDocument{\let\originallabel\label} 
\begin{document}

	\title{Open-Set Text Recognition via \Jtr{}}

	\author{Chang~Liu\\
		{\tt\small lasercat@gmx.us}
		\and
		Chun~Yang\\
		{\tt\small chunyang@ustb.edu.cn}
		\and
		Xu-Cheng~Yin\\
		{\tt\small xuchengyin@ustb.edu.cn}
	}
	
	\affiliation{Institution1}
	\author{Chang~Liu$^1$ \quad Chun~Yang$^{1,*}$ \quad Xu-Cheng~Yin$^{1,2,}$\thanks{Corresponding authors.}\\
		{\tt\small $^1$ School of Computer and Communication Engineering, University of Science and Technology Beijing}\\
		{\tt\small $^2$ Institute of Artificial Intelligence, University of Science and Technology Beijing, China}\\
		{\tt\small lasercat@gmx.us, \{chunyang, xuchengyin\}@ustb.edu.cn}\\ 
	}
	
	\maketitle
\begin{abstract}
	\chgclarify{
		The open-set text recognition task is an emerging challenge that requires an extra capability to cognize novel characters during evaluation. We argue that a major cause of the limited performance for current methods is the confounding effect of contextual information over the visual information of individual characters.  Under open-set scenarios, the intractable bias in contextual information can be passed down to visual information, consequently impairing the classification performance. In this paper, a Character-Context Decoupling framework is proposed to alleviate this problem by separating contextual information and character-visual information. Contextual information can be decomposed into \tti{} and \tli{}. Here, \tti{} that models character order and word length
		is isolated with a detached temporal attention module.  \Tli{} that models n-gram and other linguistic statistics is separated with a decoupled context anchor mechanism.} A variety of quantitative and qualitative experiments show that our method achieves promising performance on open-set, zero-shot, and close-set text recognition datasets.
\end{abstract}
	%
	%
	%
	%
	\section{Introduction}

Text recognition is a well-studied task and has been widely applied in various \chgrevD{applications~\cite{ocrsurvey}.} Most existing text recognition methods assume characters in the testing set are covered by the training set. \chgrevA{Moreover, consistency of contextual information between the training set and the testing set is also assumed.}  
These methods are not adaptable to recognize unseen characters without retraining the model. However, as the language evolves, \chgsevC{novel ligatures (e.g., rare characters, emoticons, and foreign characters)} can be frequently used in a region during a certain period. For example, foreign characters can be seen frequently in scene text images as a result of globalization. Hence, it is unfeasible if the model needs to be retrained whenever a ``new character'' emerges. \chgrevD{This} task is defined as the open-set text recognition task~\cite{neko20nocr}, as a specific field of open-set recognition~\cite{tosr} and a typical case of robust pattern recognition~\cite{ZhangLS20}. %
\chgrevA{Currently, a few visual-matching-based text recognition methods are capable to recognize novel characters in text lines~\cite{eccv20,neko20nocr,jinic21}.}

\begin{figure}[t]
	\centering
	\includegraphics[width=0.8\linewidth]{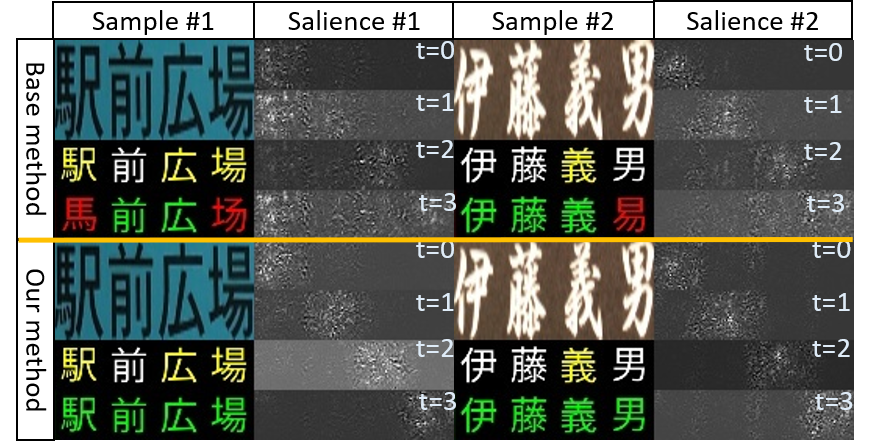}
	\caption{\chgrevA{Illustration of the ``salience region''~\cite{deepinside} of each timestamp, showing where the models look at. Base model (top) tends to seek the help from the context, while our framework (bottom) focuses more on local character features.} }
	\label{fig:cof}
\end{figure}

However, \chgclarify{these} open-set text recognition methods tend to be affected by \tsi{} captured from the training set. \chgrevA{This phenomenon can be seen in the salience map (Fig.~\ref{fig:cof})~\footnote{\url{https://github.com/MisaOgura/flashtorch}}, and is also observed in~\cite{tvr}}. In such cases, feature representation for each character is always mixed with \tli{}. 
This could benefit close-set scenarios \chgclarify{where the \tsi{} bias between training and evaluation is negligible,}
as some characters (e.g. `0' and `O') are hard to separate only by \tvi{} \chgclarify{(glyph shapes)}. However, under open-set scenarios, \tsi{} could be severely biased from the training set. \chgclarify{Consequentially, existing} model\chgclarify{s} may mistakenly ``correct'' a  character into a
\chgclarify{wrong one that fits ``better'' in the context according to the} training set~\cite{tvr}.

\chgclarify{To alleviate the impact of \tsi{} over open-set text recognition, we propose a \jtrf{}} allowing explicit separation of \tvi{} and \tsi{}. \chgclarify{\Tsi{} is further
	decomposed} into \tti{} and \tli{}.  \chgclarify{In general, \tti{}} models the number and \chgclarify{order} of characters in a word, \chgclarify{while \tli{}} models n-gram and other linguistic statistics.  \chgclarify{Accordingly,}  a \Gta{} module (DTA) is introduced to model \tti{} and isolate it from visual features.
Also, a \Ca{} mechanism (DCA) is proposed to ``explain away~\cite{expa}'' the \tli{}  from \tvi{}. \chgclarify{In summary, our framework reduces the confounding effect of training-set \tsi{} on visual features, making it less vulnerable to the intractable \tsi{} bias under open-set scenarios.} 

The main contributions of this paper are summarized as follows: 

(1) Proposing a \JtrF{} that improves word-level open-set text recognition by reducing the effect of contextual information on the visual representation of \chgclarify{novel} characters in word-level samples.

(2) Proposing a \Gta{}  module that reduces the impact of temporal information over the visual feature \chgclarify{extractor.}

(3) Proposing a \Ca{} mechanism that enables \chgsevC{the} separation of linguistic information  \chgclarify{from the visual feature extractor.}

	\section{Related Work}
	Open-set text recognition, as a specific field of open-set recognition~\cite{tosr,osr-survey}, is a task that requires the model to recognize \chgclarify{testing-set} words that may contain novel characters unseen in the training set~\cite{neko20nocr}. A few methods~\cite{eccv20,neko20nocr} have been proposed to address this task. Wan~\etal~\cite{eccv20} proposed to match the visual features of the word image with the glyph image, and \chgclarify{to} sample the matching results with a class aggregator. Their method does not scale well on large-scale character sets due to the size growth of the glyph images and the similarity maps. 
	On the other hand, OSOCR~\cite{neko20nocr} generates class centers from \chgclarify{individual glyphs} with a ProtoCNN and matches the class centers with serialized visual features of the word image. The character-based prototype generating design allows \chgclarify{reducing the training cost by mini-batching} the label set, \chgclarify{thus can be applied to larger label-sets.} However,  \chgrevA{these methods~\cite{neko20nocr, eccv20} do not provide effective approaches to separate \tsi{}, limiting} the performance of open-set word-level \chgfiveC{recognition}. \chgclarify{Impacts of \tsi{}} are also studied in~\cite{tvr}, which suggests that RNN-free methods are also prone \chgclarify{to \tsi{} bias.} \chgrevA{Hence, we propose a framework that decouples and isolates  \tsi{} from \tvi{} to improve open-set visual-matching accuracy. }
	
	The conventional close-set text recognition \chgclarify{tasks} can be considered as a special case where the testing set has \chgclarify{zero} novel characters. \chgninC{In most} conventional text recognition methods~\cite{Rosetta,SAR,CRNN,DAN,ace}, class centers are mostly modeled as weights in linear classifiers\chgclarify{, while} visual information and contextual information are modeled together \chgninC{without explicit separation}.  Recently, more methods opt to adopt dedicated post-processing fashioned modules~\cite{paddle,dsm} to model contextual information. %

	The zero-shot character recognition task is another special case of open-set text recognition. \chgclarify{Many} methods~\cite{hde,fewran,dran,taktak,jinic21} propose to encode each character with a unique structural representation \chgclarify{(e.g., radical or stroke sequences)} for prediction.  Recently, \chgclarify{a few} methods demonstrate capabilities for  Korean character recognition~\cite{taktak} and whole word recognition~\cite{jinic21}. Despite \chgclarify{performing reasonably well with large label sets, these methods require language-specific structural representations of characters, thus limit them to corresponding languages.}  In contrast, structure-free methods like~\cite{cm19,neko20nocr} only require a template from a font (or a printed \chgclarify{sample}) \chgclarify{for each character}. This approach benefits scenarios where little prior knowledge of character composition can be given, for example, ancient writings of the Oracle characters. Our method follows the structure-free scheme and further achieves reliable word recognition capability by introducing \jtr{}.
	\begin{figure*}[!ht]
		\centering
		\includegraphics[width=\linewidth]{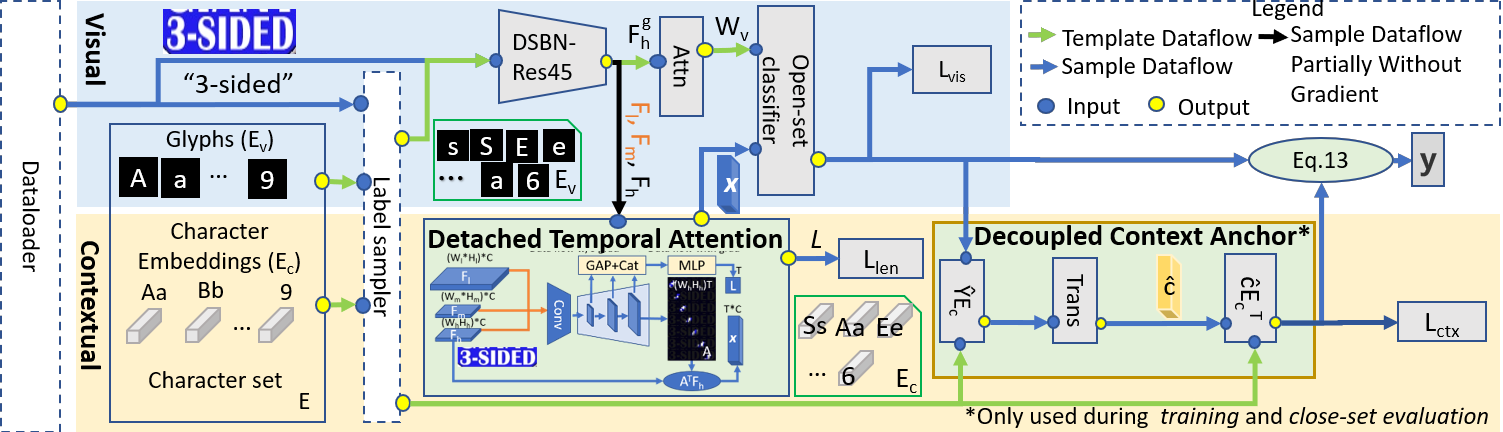}
		\caption{Our implementation of the proposed \JtrF{}.  \chgeigC{In the framework, visual representation of the sample and character templates are first extracted with the DSBN-Res45 Network~\cite{DSBN}, then the \Gta{} module predicts the word length and samples visual features $\imgat{t}$ for each timestamp. The visual prediction is achieved by matching prototypes (attention-reduced template features) with the Open-set classifier. Finally, the visual prediction is adjusted with the \Ca{} module, and no adjustment is conducted when there is intractable~\tli{} under open-set scenarios.}}
		\label{fig:jtrf}
	\end{figure*}
	
	\section{Proposed Method}
	In this work, we propose a \jtrf{} (shown in Fig. \ref{fig:jtrf}) to reduce \chgclarify{the impact of} \tsi{} bias under open-set scenarios, by separating and isolating \tvi{} and \tsi{} with the \Gta{} module and the \Ca{} mechanism. The framework and its optimization are first formulated in Section \ref{ch:fr}. Then, a detailed explanation \chgclarify{of} the less intuitive \Ca{} mechanism \chgclarify{is} presented in Section \ref{ch:ca}. Finally, the \Jtrn{} (OpenCCD) is given as an example implementation of our framework in Section \ref{ch:gta}.
	
	\subsection{\JtrF}
	\label{ch:fr}
	
	The framework takes a sample \chgclarify{(word-level image)} $img$ and \chgclarify{a} character set \chgclarify{$\templates{}$} as input, and outputs the \chgclarify{predicted} word $\predseq:(\predat{0},...,\predat{t})$ with the maximum probability given the \chgclarify{sample} and \chgclarify{character set,} 
	\begin{equation}
		\begin{split}
			\predseq=\arg \max_{\anyseq}P(\anyseq|\imgword,\templates;\theta),
		\end{split}
	\end{equation}
	\chgeigC{where $\imgword$ is the \chgclarify{visual} feature representation  \chgclarify{of all characters in the sample}.}
	We omit the $\templates$ and $\theta$ in the following part for \chgrevA{writing convenience}. In our framework, we expand $P(\anyseq|\imgword)$ with \chgsevC{a predicted length $l$, using the law of total probability},  
	\begin{equation}
		\begin{split}
			P(\anyseq|\imgword)=\sum_{l=1}^{maxL} P(l|\imgword)P(\anyseq|\imgword,l),\\
		\end{split}
	\end{equation}
	where $maxL$ is the maximum length of a word. 
	\chgclarify{Different from most existing text recognition frameworks using end-of-speech~\cite{DAN,ASTER}, segmentation~\cite{cafcn,ts}, or blanks~\cite{aon,CRNN} to handle lengths, our framework explicitly predicts the length.} 
	$P(\anyseq|\imgword,l)$ can be further decomposed to contextual predict\chgclarify{ion} and visual prediction \chgsevC{via the proposed \Ca{} mechanism (detailed in Section \ref{ch:ca})}, 
	\begin{equation}
		\begin{split}
			&P(\anyseq|\imgword,l)\\
			=&\dcaright.\\
		\end{split}
	\end{equation}
	\chgsevC{Here}, $\ctxat{t}$ is the \chgclarify{common ``context'' (\tli{})} of characters,  $\imgword$ models  the \chgclarify{visual information of all characters in the input image}, and $\imgat{t}$ corresponds to \chgclarify{the \tvi{}} of the $t^{th}$ character.
	Hence, the optimization goal would be maximizing the log-likelihood $logP(\gtseq|\imgword)$ of the \chgclarify{ground truth} label sequence $\gtseq$,
	\begin{equation}
		\begin{split}
			\label{lenloss}
			&logP(\gtseq|\imgword)\\
			=&log(\sum_{l=1}^{maxL} P(l|\imgword)P(\gtseq|\imgword,l))\\
			\labelrel={lenloss:1}&logP(l^{*}|\imgword)+ logP(\gtseq|\imgword,l^{*})\\
			=&logP(l^{*}|\imgword)+\sum_{t=1}^{l^{*}}(logP(\gtat{t}|\imgat{t}))\\
			+&\sum_{t=1}^{l^{*}}log(\anchoreddist{P(\gtat{t}|\ctxat{t})P(\ctxat{t}|\imgword,l^{*}))}\\
			:=&-(L_{len}+L_{vis}+L_{ctx}),
		\end{split}
	\end{equation}
	\chgrevA{where  $L_{len}$, $L_{vis}$, and $L_{ctx}$ are the corresponding cross-entropy losses of the three loglikelihood terms.} Step~\eqref{lenloss:1} \chgclarify{holds because the} correct label can only be predicted when the length is correctly predicted.  %
	\begin{figure}[t]
		\centering
		\includegraphics[width=\linewidth]{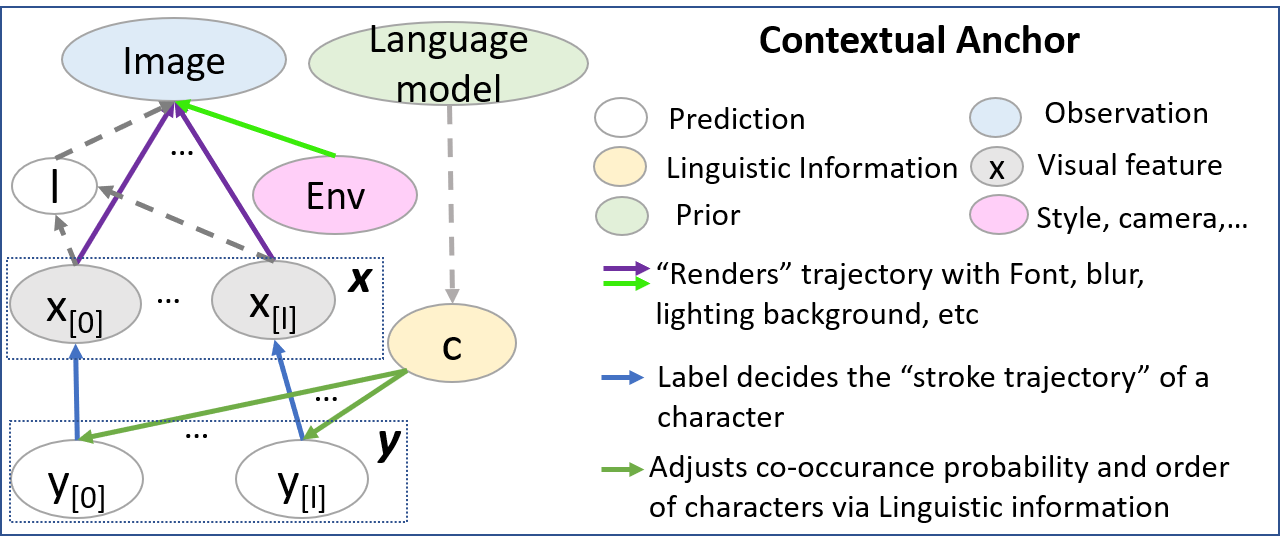}
		\caption{\chgninC{Causal graph of our Decoupled Context Anchor mechanism}. 
		}
		\label{fig:cau}
	\end{figure}
	
	\subsection{\Ca{} Mechanism}
	\label{ch:ca}
	In this work, we propose a \Ca{} mechanism to model and separate the effect of \tli{} $\ctxat{t}$ over character $\anypredat{t}$ at each timestamp $t$. 
	\paragraph{Assumption 1 (A1)}
	We assume the \tli{} functions as a common-cause of the input visual information and the prediction outputs \chgclarify{at all timestamps} (See Fig. \ref{fig:cau}).
\chgeigC{We model the sample image as a ``rendered'' result of the label $\anyseq$. Also, we \chgclarify{assume} the label (words) is generated according to \tli{} $c$, making the \chgclarify{label}  $\anyseq$ a causal result of $c$.}  Hence, linguistic context $\ctxat{t}$ and the character-level visual information $\imgat{t}$ are the only two direct factors affecting the probability of $\anypredat{t}$ at timestamp $t$, 	\begin{equation}
		\label{eq:aa1}
		P(\anypredat{t}|\imgat{t},\imgword,\anypredat{t-1}...\anypredat{0},l,\ctxat{t})=P(\anypredat{t}|\imgat{t},\ctxat{t}), 
	\end{equation}
	
	\paragraph {Assumption 2 (A2)}
The shape \chgclarify{(\tvi{})} of a character and its context \chgclarify{(\tli{})} are independent given the character \chgclarify{$y_{[t]}$}, i.e.,  	\begin{equation}
		\label{eq:aa2}
		\begin{split}
			P(\imgchanya|\predchanya,\ctxchanya)&=P(\imgchanya|\predchanya)\\
			\iff P(\imgchanya,\ctxchanya|\predchanya)&=P(\imgchanya|\predchanya)P(\ctxchanya|\predchanya).
		\end{split}
	\end{equation}
	This assumption implies that the \tli{} does not affect the ``style'' \chgclarify{(font face, color, background, etc.)} of the word, which generally holds in most synthetic datasets where  \chgclarify{styles} and \chgclarify{contents} are randomly matched.

	\paragraph{Theorem 1: The Anchor Property of Context\newline}
	
	\textit{\chgclarify{Given assumption A1}, the probability of a predicted word $\anypredseq$ given image $\imgword$ and its length $l$, $P(\anypredseq|\imgword,l)$, can be written as the product of the ``anchored prediction\chgclarify{s}'' of all timestamps, i.e., }
	\begin{equation}
		\begin{split}\label{canc}
			&P(\anypredseq|\imgword,l)=\prod_{t=1}^l \anchoreddist{P(\anypredat{t}|\imgat{t},\ctxat{t})}P(\ctxat{t}|\imgword,l),\\
		\end{split}
	\end{equation}
	\chgclarify{and the proof is detailed in Appendix~\ref{prof:1}.} Here, the integral term can be interpreted as an ensemble of ``anchored prediction'' $P(\anypredat{t}|\imgat{t},\ctxat{t})$ over all possible contexts $c$, which is similar to the hidden anchor mechanism~\cite{ham}. Hence, we call this theorem the anchor property of context.  

	\paragraph{Theorem 2: The Separable  Property of \TlI{} and \TvI{} \newline }
	\textit{Given Assumption A2, the effect of \tvi{} over the label $P(\predchanya|\imgchanya)$ and the effect of \tli{} $P(\predchanya|\ctxat{t})$ \chgclarify{is separable} from contextual prediction  $P(\predchanya|\imgchanya,\ctxchanya)$},
	\begin{equation}
		\begin{split}
			P(\predchanya|\imgchanya,\ctxchanya)\propto\frac{P(\predchanya|\imgchanya)P(\predchanya|\ctxchanya)}{P(\predchanya)}.\\
		\end{split}
	\end{equation}
	Here, $P(\predchanya|\imgchanya)$ represents the \chgclarify{predicted probability} of $\predchanya$ with regard to \tvi{} $\imgchanya$, $P(\predchanya|\ctxchanya)$ models the effect caused by  \tli{}, and $P(\predchanya)$ models the character frequency on the training set. The proof of this theorem is given in Appendix \ref{prof:2}. This theorem suggests that the effect of \tvi{} and \tli{} over \chgclarify{the} prediction can be \textbf{explicitly} separated \chgclarify{under} specific conditions.
	
	Intuitively,  $P(\predchanya|\ctxchanya)$ ``explains away''~\cite{expa} the \tli{} from the visual-based prediction $P(\predchanya|\imgchanya)$. This behavior happens in the backpropagation pass of our framework during training, where the gradients of $L_{ctx}$ and $L_{vis}$ are accumulated to update the feature extractor. This is  \chgclarify{the reason that  $L_{ctx}$ needs to be} backpropagated\chgrevD{, and also makes $L_{ctx}$ a regularization term,  in terms of enforcing certain properties  of the network via backpropagation.  This property  differentiates it from the ``look-twice'' mechanisms~\cite{dsm,paddle} that cut gradients.}%

	\paragraph{\chgclarify{Theorem 3:} \Ca{} Mechanism \newline}
	Combining Theorem 1 and Theorem 2, we have the \Ca{} mechanism,
	\begin{equation}\label{thr}
		\begin{split}
			&P(\anypredseq|\imgword,l)\\
			=&\dcaright.\\
		\end{split}
	\end{equation}
	\chgclarify{Proof of this theorem can be found in Appendix~\ref{prof:3}. }
	The mechanism further allows explicit separation of \tli{} and \tvi{} on the \textbf{word} level, which provides a way to model and separate \tli{} learned on the training set, resulting in a feature extractor \chgclarify{
		focusing more on  \tvi{} and less affected by the training set \tli{}.}  Considering the anchor property revealed in Theorem 1 and the decoupling 
	\chgclarify{nature of Theorem 2, }%
	we call this mechanism \chgfiveC{the} \Ca{} \chgclarify{mechanism}.  

	\subsection{\chgclarify{OpenCCD Network}}
\chgclarify{In this section,} the Open-set Character-Context Decoupling network  (OpenCCD, Fig. 2) is given as an example implementation of our proposed framework. \chgclarify{Here, character set  $E:(E_{v},E_{c})$ consists of glyphs from the Noto font $E_v$ and semantic embeddings of the characters $E_c$.} 
The \chgclarify{network} first extracts visual features of  the word images \chgclarify{$img$} and the glyphs $E_{v}$ with the \chgclarify{45-layer ResNet built with DSBN~\cite{DSBN} layers (Res45-DSBN). It shares the convolutional layers between the glyphs and word images, while keeping task-specific batch statistics. Three levels of word features $(F_l,F_m,F_h)$ and the latest feature map $F^{g}_h$ of glyphs are used.} The prototypes  (class centers) $W_{v}$ are generated by applying geometric attention \chgclarify{to $F^{g}_h$}. During training, we mini-batch  $E_{v}$ \chgclarify{at each iteration} to achieve \chgclarify{a} reasonable training speed. During the evaluation, the visual prototypes $W_{v}$ for the whole dataset are cached beforehand, \chgclarify{hence prototype generation yields little extra costs.}

Next, the \Gta{} (DTA) module is used to predict the length of the word $P(l|\imgword)$, \chgclarify{ and the max probable length is detonated as $\hat{l}$.} Then \chgclarify{the DTA module} samples ordered character-level visual features  $\imgword:(\imgat{0},...,\imgat{\hat{l}})$ from the feature map $F_h$.  

\chgclarify{The visual-based prediction $P(\anypredat{t}|\imgat{t})$ is then produced by the open-set classifier.  
	For close-set scenarios, the \tli{} oriented prediction $P(\anypredat{t}|\ctxat{t})$  is produced via the \Ca{}~(DCA) module.} \chgsevC{For open-set scenarios where \tli{} is intractable, \chgclarify{$P(\anypredat{t}|\ctxat{t})$} is treated as a uniform distribution\chgclarify{, which is} equivalent to \chgclarify{only} using the visual prediction.}

\begin{figure}[]
	\centering
	\includegraphics[width=0.85\linewidth]{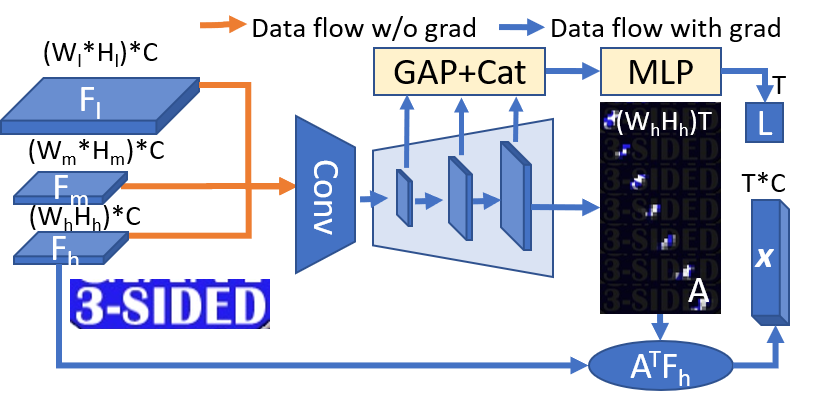}
	\caption{The proposed \gta{} module. We isolate sequence modeling within the Temporal Attention module, and zero the gradient of convolution features, w.r.t., the temporal attention map. \chgeigC{Here, GAP indicates a global-average pooling.}}
	\label{fig:gta}
\end{figure}

\paragraph{\Gta{}}
\label{ch:gta}

\chgrevD{
	In OpenCCD, the \gta{} module (Fig. \ref{fig:gta}) is proposed to predict the sequence length $P(l|\imgword)$. It also sort and sample character in feature map $F_h$ via the attention map $A$. The module utilizes an FPN to model global \tti{} from the input feature maps, and decodes them into $A$ and $P(l|\imgword)$. Since \tti{} is  not related to individual character shapes (\tvi{}), the module novelly isolates it from input visual feature maps by cutting the gradients w.r.t. $P(l|\imgword)$ and $A$. 
	The module then segments input visual feature map $F_h$ into visual features  $\imgword$ for individual characters according to attention map $A$ and the most probable length $\hat{l}$, allowing only \tvi{} backpropagating to $F_h$ via $\imgword$. 
}  %
	In \JTRN{}, $P(\anypredat{t}|\imgat{t})$ is produced by comparing the prototypes with character-level feature\chgclarify{s} $\imgat{t}$,
	\begin{equation}
		\label{eq:rej}
		P(\anypredat{t}|\imgat{t})\propto 
		\begin{cases}
			\alpha|\imgat{t}| &\anypredat{t}\text{is [UNK]}\\
			|\imgat{t}|Sim(\imgat{t},\anypredat{t})& \text{otherwise}, \\
		\end{cases}
	\end{equation}
	where $|\imgat{t}|$ is the \chgclarify{L2-N}orm of $\imgat{t}$,``[UNK]'' indicates unknown characters, and $\alpha$ is a trainable similarity threshold for rejection.  $Sim(\imgat{t},\anypredat{t})$ is defined as
	\begin{equation}
		Sim(\imgat{t},\anypredat{t}):=\max_{w_{v}\in \psi(\anypredat{t})}(cos(w_{v},\imgat{t})),
	\end{equation}
	where $\psi$ returns all prototypes \chgclarify{$\psi(\anypredat{t}) \subset  W_{v}$ associated with} label $\anypredat{t}$, \chgclarify{and each individual prototype $w_{v}$} corresponds to a ``case'' of character $\anypredat{t}$. 
	\paragraph{\Ca}
	Instead of implementing a Variational Auto Encoder~\cite{vae} to estimate the distribution of \chgclarify{\tli{}} and estimate the integral with Monte-Carlo, we approximate the integral with predicted context $\hat{c}$, \chgrevD{which is similar to} the conventional anchor mechanisms \chgclarify{using only} the anchor with maximum prediction likelihood~\cite{faster,yolo2},
	\begin{equation}\label{eq:aproxim}
		\begin{split}
			\anchoreddist{P(\anypredat{t}|\ctxchanya) P(\ctxat{t}|\imgword,l)}
			\approx&P(\anypredat{t}|\ctxestat{}).\\
		\end{split}	
	\end{equation}
	\chgclarify{Combined with Eq.~\ref{thr}}, the probability of a predicted character at \chgclarify{timestamp} $t$ can be approximated as,
	\begin{equation}
		\begin{split}
			P(\anypredat{t}|\imgword) \approx P(\anypredat{t}|\imgat{t})P(\anypredat{t}|\hat{c}).
		\end{split}	
	\end{equation}

	\chgfiveC{As \tli{} is mostly related to labels, we estimate the \tli{} $\hat{c}$ from the predicted label instead of the feature map. More specifically, the module \chgclarify{reuses the estimated} character probability distribution $Y\in (0,1)^{l\times M}:(P(\anydistat{0}|\imgat{0}),...,P(\anydistat{l}|\imgat{l}))$ with regard to \tvi{} at each timestamp $t$, \chgclarify{and $P(\anydistat{t}|\imgat{t}): (P(\anypredat{t}^0|\imgat{t}),...,P(\anypredat{t}^M|\imgat{t}))$ is the probability distribution of all characters at timestamp $t$. } Then $\ctxestat{t}$ is estimated with a \chgsevC{4-layer transformer encoder \cite{aian}} applied on the expectation of character embeddings,}
	\begin{equation}
		\begin{split}
			\hat{c}=&Trans(YE_c),\\
		\end{split}
	\end{equation}
	\chgfiveC{where $E_{c} \in R^{M\times C}$ is the semantic embedding of seen characters in the training set, hence $YE_c$ interprets as expectation.  $Trans$ \chgclarify{indicates the 4-layer transformer encoder}. Finally, $P(\anydistat{t}|\ctxestat{t})$ is estimated by comparing character embedding $E_{c}$ to $\ctxestat{t}$,}
	\begin{equation}
		P(\anydistat{t}|\ctxestat{t})=\sigma(\ctxestat{}E_{c}^{T})_{[t]},\\
	\end{equation}
	where $\sigma$ is the softmax function. 
	
	\paragraph{Optimization}
	\chgsevC{With Eq.\ref{eq:aproxim} reducing the integral down to a standard classification problem, $L_{ctx}$ in Eq. \ref{lenloss} can be implemented as a cross-entropy loss like $L_{len}$ and $L_{vis}$. Hence, \JTRN{} can be optimized with the three equally-weighted cross-entropy losses. }

	\newcommand{\japit}{200k}
	
	\newcommand{\japsmean}{62.16 36.57}
	\newcommand{\japssk}{ 76.48 74.15}
	\newcommand{\japsuk}{77.50 48.81}
	\newcommand{\japsak}{77.06 59.81}
	\newcommand{\japskana}{43.55 7.52}

	\newcommand{\japlmean}{\textbf{65.34 41.31}}
	\newcommand{\japlsk}{76.53 \textbf{74.66}}
	\newcommand{\japluk}{\textbf{82.20 58.33}}
	\newcommand{\japlak}{\textbf{79.74 65.42}}
	\newcommand{\japlkana}{\textbf{47.35 11.17}}
	
	\newcommand{\japfigperf}{figure-200k/samplejapva}
	\newcommand{\japfigabl}{figure-200k/abl-lat-tall-2}
	\newcommand{\japfigablq}{figure-200k/abl-oss}

	\newcommand{\chfigperf}{figures/remixch2}
	
	\newcommand{\chperf}{Ours~ &-&\textbf{90.93} & \textbf{94.10}& \textbf{94.58}& \textbf{95.55} &\textbf{58.22} &\textbf{68.56} & \textbf{74.45} & \textbf{77.18}}

	\newcommand{\mjstfpslap}{66.91}
	\newcommand{\mjstfpslapm}{255}
	\newcommand{\mjstfpsserm}{327}
	\newcommand{\maxram}{2.5}

	\newcommand{\mjstitr}{800k}
	\newcommand{\mjstperf}{\mjstfpslap{}/\textbf{\mjstfpslapm}$^{+}$&91.90&85.93&92.38&92.21&83.68}
	
	\newcommand{\mjstdperf}{\textbf{99.8}/\textbf{99.0}&\textbf{96.9}&\textbf{97.9}}
	
	\newcommand{\tvdta}{2.00}
	\newcommand{\pvdta}{0.06}
	\newcommand{\pvfull}{1.54\times 10^{-5}}
	\newcommand{\tvfull}{5.87}
	
	\section{Experiments}
	The work is based on the OSOCR~\cite{neko20nocr}, our code\footnote{https://github.com/lancercat/VSDF} and datasets\footnote{https://www.kaggle.com/vsdf2898kaggle/osocrtraining} are released. We conduct experiments on benchmarks for all three scenarios: open-set word-level recognition, \chgeigC{zero-shot character} recognition, and the conventional close-set word-level recognition benchmarks. Moreover, the \chgclarify{ablative} studies for open-set word-level recognition \chgclarify{are also} performed.  We use the AdaDelta optimizer, the learning rate is set to $10^{-2}$, and decreases by \chgclarify{every} $200k$ iterations. \chgrevD{For word recognition tasks, we provide a ``large'' network for an alternative speed-performance trade-off profile closer to SOTA methods, where the large network has more latent channels in the ResNet45-DSBN backbone.}%

	\subsection{Open-Set Text Recognition}
\begin{figure}
		\centering
		\includegraphics[width=0.9\linewidth]{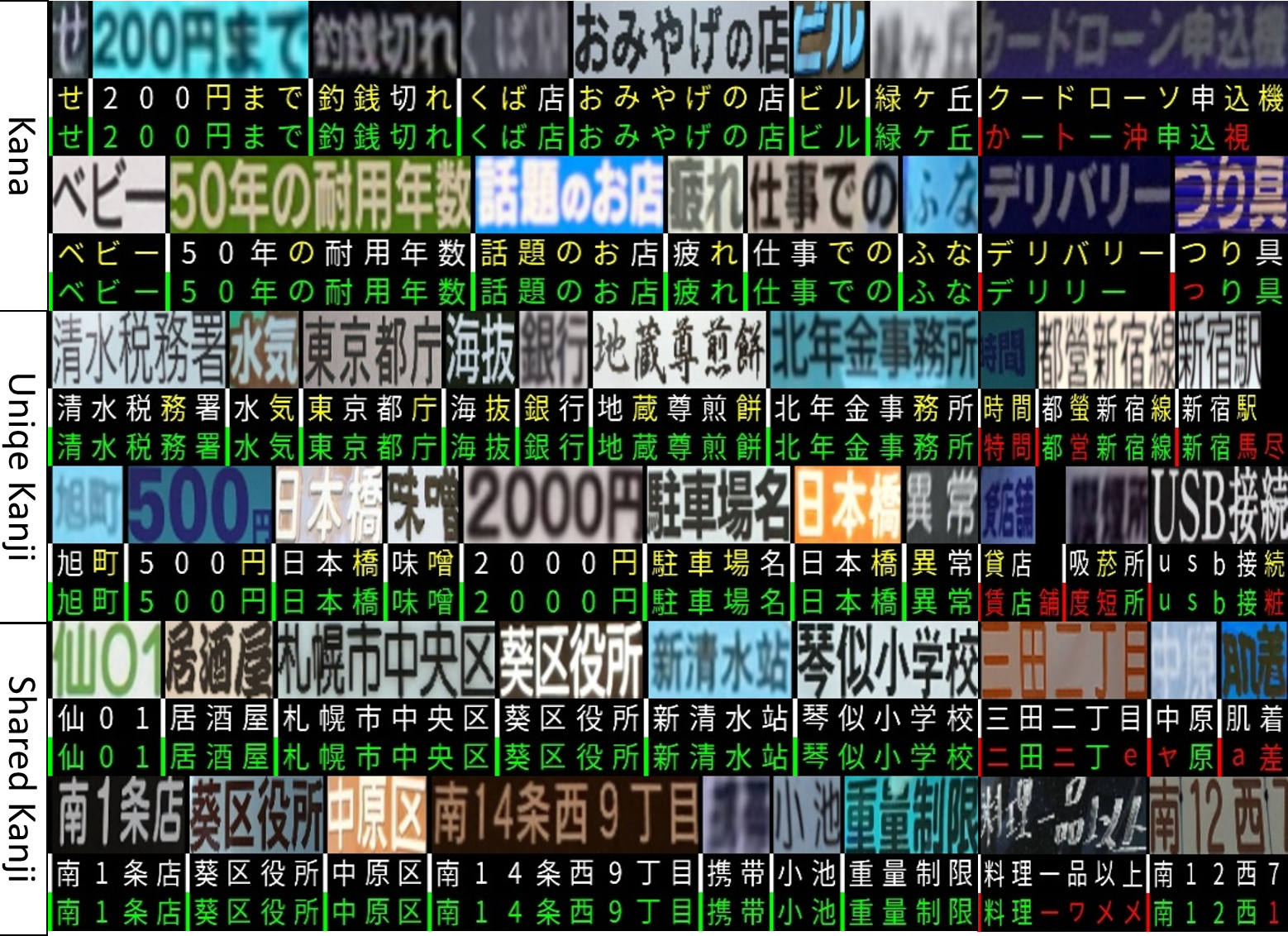}
		\caption{
			Sample results from the open-set text recognition task. The figure shows qualitative performance under the ``Kana'', ``Unique Kanji'' and ``Shared Kanji (close-set)'' scenarios. The results for each group are represented with two rows, where the top row shows the success cases and the bottom shows failure cases. Text in white indicates seen characters, yellow indicates novel characters, red indicates recognition error, green indicates correct results, and purple block indicates rejected results.
		}
		\label{fig:goodNbad}
	\end{figure}

	We use a collection of Chinese text recognition dataset\chgclarify{s} as the training set and the Japanese subset of MLT as the testing set following OSOCR~\cite{neko20nocr}, and all models are trained for \japit{} iterations.  %
	\chgclarify{Quantitative performances are shown in Table ~\ref{tab:perfOST} along with SOTA methods}, and qualitative samples can be found in Fig \ref{fig:goodNbad}. 
	The results show overall significant performance improvement over OSOCR~\cite{neko20nocr}. Details suggest the performance gain comes from recognizing unseen characters. 
	The model shows some extent of robustness over novel characters (text with yellow color in Fig\chgclarify{.} \ref{fig:goodNbad}) like unique Kanjis and Kanas.  
	\begin{table}[!t] 
		\caption{
			Detailed performance analysis on the open-set text recognition dataset. Performance data listed in Character Accuracy~(\textbf{top}) / Line Accuracy (\textbf{bottom}) manner.  
		}
		\begin{center}
			\centering
			\begin{tabular}{P{1.3cm}|P{1.3cm}|P{1.3cm}|P{1.3cm}|P{1.3cm}}
				\hline
				Method       & OSOCR \cite{neko20nocr}       & OSOCR Large~\cite{neko20nocr}  & Ours        & Ours Large  \\
				\hline
				Kana         & -           & 18.75 0.10  & \japskana{}  & \japlkana{}\\
				\hline
				Shared Kanji & -           & \textbf{79.86} 73.81 &\japssk{} & \japlsk{} \\
				\hline 
				Unique Kanji & -           & 71.74 34.08 & \japsuk{} & \japluk{} \\
				\hline
				All Kanji    & -           & 75.33 51.64 & \japsak{} & \japlak{} \\
				\hline
				Overall      & 47.89 29.08 & 49.10 30.08 & \japsmean{} & \japlmean{} \\
				\hline
				
			\end{tabular}
			\label{tab:perfOST}
		\end{center}
	\end{table}
	
	\chgclarify{Results} indicate that characters having close shapes are the most significant source of mistakes. The reason for this phenomenon could be pushing all negative \chgclarify{classes} alike with hard labels \chgclarify{(in contrast to soft-label)}, which is also mentioned in fine-grain classification~\cite{soft}. 
	Blur, text art can be another major cause for the failure \chgclarify{cases}, \chgsevC{which is expectable} as  \tli{} is intractable under open-set scenarios, \chgclarify{consequentially cannot be used to recover the visually indistinguishable characters.}

	\subsection{\chgfiveC{Ablative Study}}
\begin{figure}[t]
		\includegraphics[width=\linewidth]{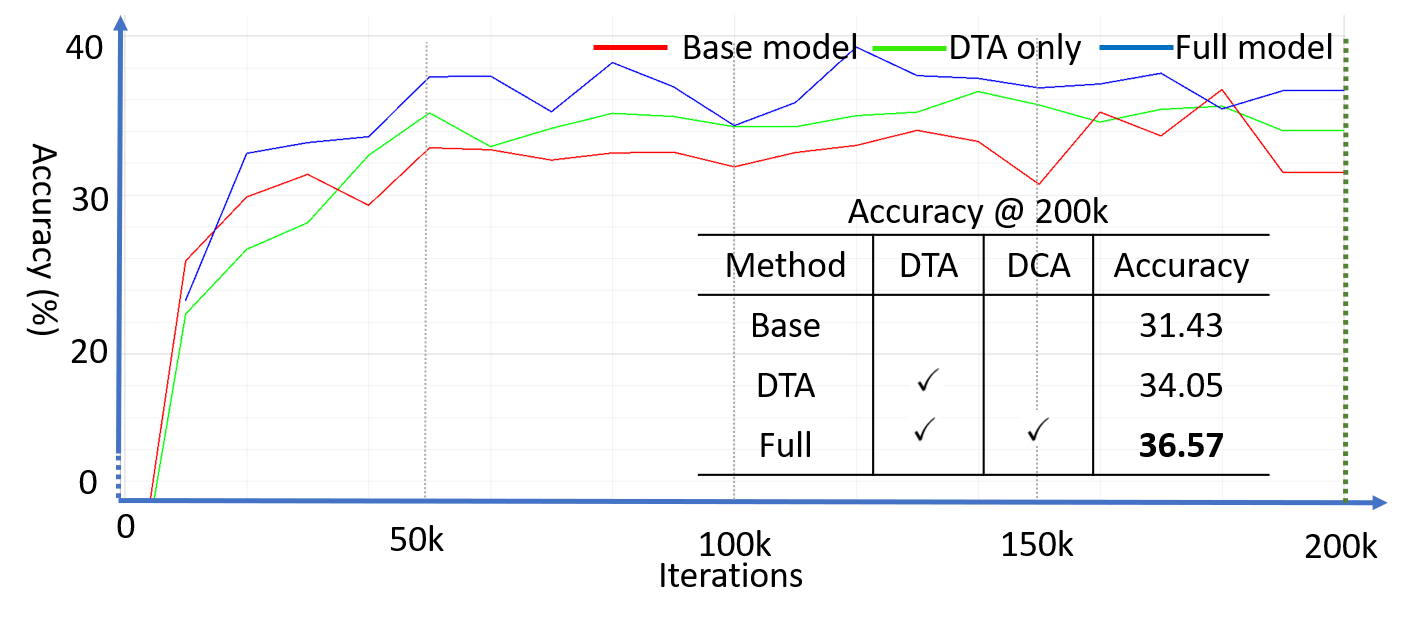}
		\caption{Ablative study on modules proposed. X-axis indicates total iteration and Y-axis indicates test-set accuracy. Steady-preformance gain can be achieved after applying each module (Red: base model; green: detaching \tvi{}; blue: detaching both).}
		\label{fig:abl}
	\end{figure}

	We conduct ablative studies on the open-set text recognition challenge to validate the effect of \chgclarify{decoupling} \tvi{} and \tsi{}. In this section, we train all ablative models on the same server to minimize the confounding factors, the results \chgclarify{(Fig. \ref{fig:abl}) show that} isolating \tti{} with the \Gta{} module can improve the open-set recognition performance.  Also, further separating  \tli{} with the \Ca{} mechanism is proved to yield more improvement. 
	\begin{table*}[!t]
		\caption{
			Zero-shot character recognition accuracy on HWDB and CTW datasets. * indicates ``online trajectory''  data required.
		}
		\begin{center}
			\begin{tabular}{c|c|c|c|c|c|c|c|c|c}
				\hline
				& & \multicolumn{8}{c}{Accuracy (\%)} \\
				\cline{3-10}
				& &\multicolumn{4}{c|}{HWDB}&\multicolumn{4}{c}{CTW}\\
				\cline{3-10}
				Method&Venue&\multicolumn{4}{c|}{\# characters in training set}&\multicolumn{4}{c}{\# characters in training set} \\
				\cline{3-10}
				& &500& 1000&1500 &2000& 500& 1000&1500 &2000 \\ 
				\hline
				CM*~\cite{cm19}&ICDAR'19&44.68&71.01&80.49&86.73&-&-&-&-\\
				\hline	
				\hline	
				DenseRan~\cite{denseran}& ICFHR'18 &1.70&8.44&14.71&63.8&0.12&1.50&4.95&10.08\\
				\hline
				FewRan~\cite{fewran}& PRL'19&33.6&41.5&63.8&70.6&2.36&10.49&16.59&22.03\\
				\hline
				HCCR~\cite{hde}&PR'20&33.71&53.91&66.27&73.42&23.53&38.47&44.17&49.79\\
				\hline
				OSOCR~\cite{neko20nocr}&-&46.67&72.19&79.82&84.31&27.94&48.23&58.56&63.77\\
				\hline	
				\chperf{}\\
				\hline	
			\end{tabular}
			\label{tab:perfch}
		\end{center}	
	\end{table*}	
	
	\chgclarify{Intuitively,} test-set accuracy curves show that both \chgclarify{proposed} approaches introduce a steady performance improvement \chgclarify{on most iterations.}
	\chgrevD{The instability of the curves is caused by the Line Accuracy metric where one wrong character can compromise the whole line.  We further perform paired $t$-tests to quantitatively} validate the robustness of the performance improvements. Separating \tti{} with the DTA module shows a $\tvdta$ $t$-value and  $\pvdta$ $p$-value, while using  DCA to separate \tli{} gives a $\tvfull$ $t$-value and  $\pvfull$ $p$-value. \chgeigC{The $p$-values suggest we can reject the ``two-sided''(no improvement) null hypothesis for both approaches.}
	Hence, there is strong evidence that \chgrevD{both DTA and DCA can robustly improve the open-set recognition performance.}  

	\begin{figure}[!t]
		\centering
		\includegraphics[width=0.8\linewidth]{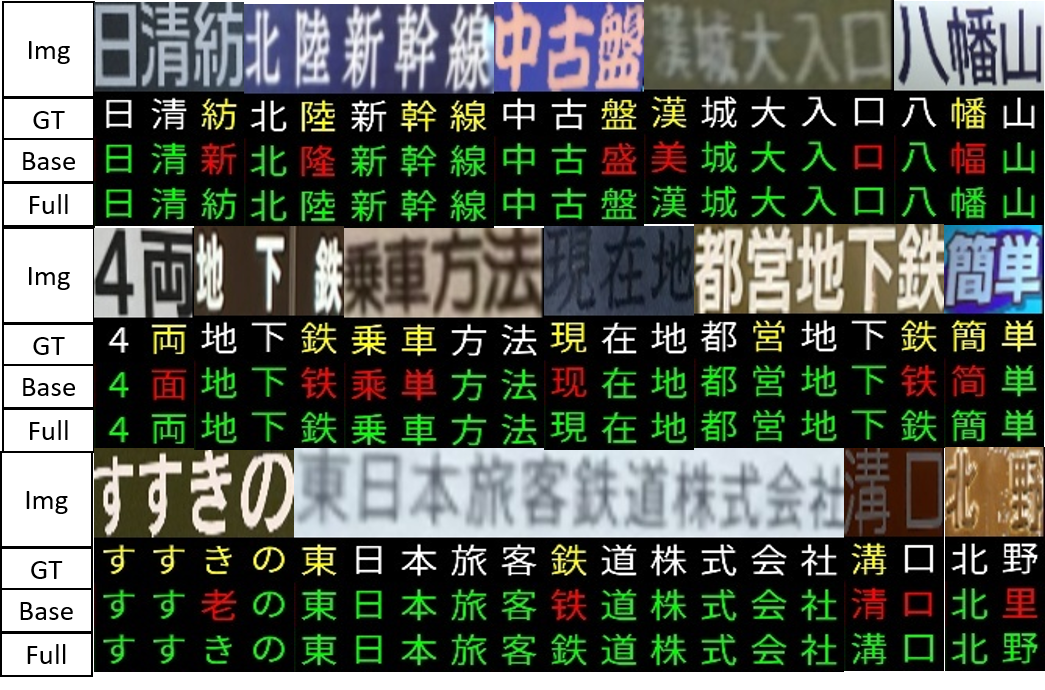}
		\caption{\chgeigC{Comparison between our method and the base method. Green indicates correct predictions, red indicates wrong predictions. Yellow indicates novel characters and white indicates seen characters. }}
		\label{fig:abl-x}
	\end{figure}
	\chgeigC{
		Qualitatively, we show the result comparison between our model and the base model in Fig. \ref{fig:abl-x}. Our framework demonstrates decent robustness improvement against the linguist information bias compared to the base model by separating \tli{} and \tvi{}.%
	}

	\subsection{Conventional Benchmarks}
	\chgrevD{
		Due to the lack of open-set text recognition benchmarks, we adopt two well-studied special cases to give referenced comparisons on generalization capability and word recognition capability.  Here, we stick to the most applied protocols in each corresponding community to train, evaluate, and measure the performances.} 
	\begin{table*}[!t]
		\caption{
			Performance on conventional close-set benchmarks. * indicates character-level annotation and $^{+}$ for multi-batch evaluation.	}
		\begin{center}
			\begin{tabular}{c|c|c|c|c|ccccc}
				\hline
				Methods&Venue&Training Set&RNN&FPS&IIIT5K&SVT&IC03&IC13&CUTE\\
				\hline
				Comb.Best~\cite{www}&ICCV'19&MJ+ST&Y&36.23&87.9&87.5&94.4&92.3&71.8\\
				SAR~\cite{SAR}&AAAI'19&MJ+ST&Y&-&91.5&84.5&-&-& 83.3\\
				ESIR~\cite{ESIR}&CVPR'19&MJ+ST&Y&-&93.3& 90.2&-&-& 83.3\\
				SCATTER~\cite{scatter}&CVPR'20&MJ+ST+Extra&Y&-&93.7&\textit{92.7}&\textbf{96.3}&\textbf{93.9} &\textbf{87.5}\\
				SEED~\cite{seed}&CVPR'20&MJ+ST&Y&-&93.8&89.6&-&92.8& 83.6\\
				DAN~\cite{DAN}&AAAI'20&MJ+ST&Y&-&\textbf{94.3}&89.2&95.0&93.9& 84.4\\
				\hline
				Rosetta~\cite{Rosetta}\cite{www}&KDD'18&MJ+ST&N&\textbf{212.76}&84.3&84.7&92.9&89.0& 69.2\\
				CA-FCN*~\cite{cafcn}&AAAI'19&ST&N&45&92.0&82.1&-&91.4& 78.1\\  
				TextScanner*~\cite{ts}&AAAI'20&MJ+ST+Extra&N&-&93.9&90.1&-&92.9& 83.3\\
				\hline
				\textbf{Ours-Large}&-&MJ+ST&N& \mjstperf{}\\
				\hline
				
			\end{tabular}
			\label{tab:perfcl}
		\end{center}
	\end{table*}
	
	\paragraph{Zero-Shot Character Recognition}
	\label{zsocr}
	
\begin{figure}[t]
		\centering
		\includegraphics[width=\linewidth]{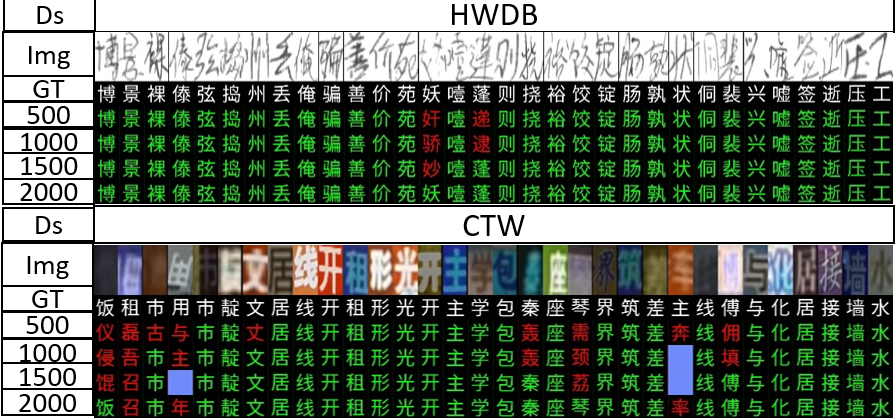}
		\caption{
			Sample results on the zero-shot Chinese \textbf{Character} recognition benchmarks. \chgninC{Numbers on the left indicates the number of different classes used for training.}  \chgeigC{ Green indicates correct recognition results, Red indicates wrong ones, and purple block indicates rejections. Note white does \text{not} indicate seen characters, \textbf{all} test characters are novel. }  
		}
		\label{fig:zerosam}
	\end{figure}

	\chgrevD{Following the common protocol in the community~\cite{hde,fewran,neko20nocr,cm19}, we perform the zero-shot Chinese character recognition benchmarks on the HWDB~\cite{hwdb} and the CTW~\cite{ctw} dataset following~\cite{hde,fewran,neko20nocr}.} The model is trained for 50k iteration\chgclarify{s} due to the small size of the training set.  As shown in Table \ref{tab:perfch}, our method shows a significant performance advantage over existing methods. \chgfiveC{Qualitative samples in Fig. \ref{fig:zerosam} show some robustness over style diversity, slight blur, and other confounding factors. \chgeigC{This also suggests that some degenerates like blur or low-contrast do not necessarily yield permanent information loss and could be inverted with sufficient well-distributed training data. 
			We owe part of the robustness difference between open-set word recognition and this challenge to the potential language-specific \chgclarify{ ``image style'' bias, caused by factors including different cameras, picture-taking habits in different corresponding region.}}}
	
	These experiments demonstrate reasonable generalization \chgclarify{capability} compared to the SOTA zero-shot character recognition methods. \chgclarify{This also justifies the choice} of the data-driven latent representation against model-driven representations like the radical \chgclarify{sequences} \cite{hde,taktak}. 
	Our method does not \chgclarify{require structural knowledge} of characters, which enables potential use to recognize Oracles and \chgclarify{other} ligatures \chgclarify{where such knowledge is unknown or not applicable}. 
	
	\label{csocr}

	\paragraph{Close-Set Benchmarks}
	Finally, we perform experiments on the conventional close-set benchmarks, where the method is compared to SOTA text recognition methods performance-wise and speed-wise. We report both dictionary-free performance (Table \ref{tab:perfcl}) and dictionary-based performance~\cite{eccv20}. More specifically, the model is trained on MJ~\cite{MJ} and ST~\cite{ST} following the mainstream technique of SOTA methods. For evaluation, IIIT5k~\cite{iiit5k}, SVT~\cite{SVT}, ICDAR 2003~\cite{ic03}, ICDAR 2013~\cite{ic13}, and CUTE~\cite{cute} are used. Our model is trained for \mjstitr{} iterations due to the significantly larger training set.

	\begin{table}[!t]
		\begin{center}
			\caption{Experiments on lexicon-based close-set benchmarks. $^c$ indicates close-set methods and $^{*}$ indicates datasets other than MJ and ST are used.}
			\begin{tabular}{p{2.3cm}|P{1.2cm}|P{1.65cm}|P{0.57cm}|P{0.56cm}}
				\hline
				Method & Venue &IIIT5k \small{(small/middle)}&IC03 (full)&SVT (50)\\
				\hline
				AON$^{c}$~\cite{aon}& CVPR'18& 99.6/98.1&96.7&96\\
				ESIR$^{c}$~\cite{ESIR}& CVPR'19& 99.6/98.8&-&97.4\\
				CA-FCN$^{c*}$~\cite{cafcn}& AAAI'19& \textbf{99.8}/\textbf{98.9}&-&\textbf{98.5}\\
				\hline
				\hline
				Zhang \etal~\cite{eccv20}& ECCV'20 & 96.2/92.8&93.3&92.4\\
				\hline
				OSOCR-L~\cite{neko20nocr}&- &99.5/98.6&96.7&96.7\\
				\hline
				Ours-L &- &\mjstdperf{}\\
				\hline
			\end{tabular}
			\label{tab:perfcld}
		\end{center}
	\end{table} 
	We first compare our method to other open-set text recognition methods that report their performances on lexicon-based benchmarks in Table \ref{tab:perfcld}, together with some popular close-set recognition methods. Results show our method retains reasonable close-set performance compared to other open-set methods. Our method \chgclarify{also} reaches close performance against SOTA close-set methods on this benchmark. 
	\chgfiveC{\chgclarify{Second,} comparisons using the dictionary-free protocol are shown in Table \ref{tab:perfcl}, despite} the performances being slightly lower than the heavy SOTA close-set recognition methods \chgrevC{to trade for faster speed}, our method shows  \chgfiveC{competitive performance} against \chgsevC{lightweight} text recognition methods.  \chgrevA{Following community convention~\cite{www,cafcn}, running speed is  adopted to measure the cost of the method.} Our method can reach 67 FPS single batched and \mjstfpslapm{} FPS multi-batched on a laptop with an RTX 2070 Mobile GPU (7 TFlops), while only using \maxram{} GiB Vram. \chgrevA{This justifies our model as a competitive light-weight method for conventional tasks.}

	\section{Limitations}
	Despite showing reasonable performance\chgclarify{s} on all tested scenarios, our method still has some limitations. Framework-wise, we made a few strong assumptions. \chgsevC{First, we assume the visual feature extractor can be generalized to a new language. Despite showing better intra-language transferring capability than radical-based methods, it is a little bit too strong to assume \chgclarify{robust} inter-language transferring capability. These limitations could be causing the performance gap between Kanas and Unique kanjis}.
	Implementation-wise, our method uses a small input ($32*128$ patches) and lacks effective rectification modules~\cite{ASTER,moran}. This leads to a very small effective text area, hence limiting the performance of skewed and curvy samples.
	We will discuss how to address these limitations in our next work.
	
	\section{Conclusion}
	In this paper, we propose a Character-Context Decoupling Framework for open-set text recognition, \chgclarify{which is theoretically sound and experimentally feasible}. Specifically, the ablative studies and comparative experiments verify that our implementation is an effective open-set text recognition method and a \chgninC{production-ready lightweight text recognition method under close-set scenarios.}%
	\section{Acknowledgement}
	The research is partly supported by the National Key Research and Development Program of China (2020AAA09701), The National Science Fund for Distinguished Young Scholars (62125601), and the National Natural Science Foundation of China (62006018, 62076024).
	\clearpage
	
	{\small
		\bibliographystyle{ieee_fullname}
		\bibliography{cvpr.bib}
	}
	\clearpage
	
	\appendix
	
	\section{Proof of Theorem 1}
	\paragraph{Assumption 1 (A1)}
	We assume the \tli{} functions as a common-cause of the input visual information and the prediction outputs \chgclarify{at all timestamps} (See Fig. \ref{fig:cau}).
\chgeigC{We model the sample image as a ``rendered'' result of the label $\anyseq$. Also, we \chgclarify{assume} the label (words) is generated according to \tli{} $c$, making the \chgclarify{label}  $\anyseq$ a causal result of $c$.}  Hence, linguistic context $\ctxat{t}$ and the character-level visual information $\imgat{t}$ are the only two direct factors affecting the probability of $\anypredat{t}$ at timestamp $t$, 	\begin{equation}
		\label{eq:aa1}
		\begin{split}
			&P(\anypredat{t}|\imgat{t},\imgword,\anypredat{t-1}...\anypredat{0},l,\ctxat{t})\\
			\labelrel{:=}{eq:aa1:1}&P(\anypredat{t}|\imgat{t},\imgword,Pre_{[t]},l,\ctxat{t})\\
			=&P(\anypredat{t}|\imgat{t},\ctxat{t}), \\
		\end{split}
	\end{equation}
	\chgsevC{where we denote prefix of $\anypredat{t}$ as $Pre_{[t]}$ in \eqref{eq:aa1:1}.}
	\begin{proof} \label{prof:1}
		\begin{equation}
			\begin{split}\label{canc}
				&P(\anypredseq|\imgword,l)\\
				=&\prod_{t=1}^lP(\anypredat{t}|\imgword,l,\anypredat{t-1}...\anypredat{0})\\
				=&\prod_{t=1}^l\anchoreddist{P(\anypredat{t}|\imgword,Pre_{[t]},l,\ctxat{t})P(\ctxat{t}|\imgword,l)}\\
				\labelrel={canc:1}&\prod_{t=1}^l \anchoreddist{P(\anypredat{t}|\imgat{t},\ctxat{t})}P(\ctxat{t}|\imgword,l).\\
			\end{split}
		\end{equation}
	\end{proof}
	Here, \eqref{canc:1} is derived by applying Eq. \ref{eq:aa1} of Assumption A1. \chgninC{The integral term can be interpreted as an ensemble of ``anchored prediction'' $P(\anypredat{t}|\imgat{t},\ctxat{t})$ over all possible contexts $c$, which is similar to the hidden anchor mechanism~\cite{ham}. Hence, we call this theorem the anchor property of context.}

	\section{Proof of Theorem 2}
	
	\paragraph {Assumption 2 (A2)}
The shape \chgclarify{(\tvi{})} of a character and its context \chgclarify{(\tli{})} are independent given the character \chgclarify{$y_{[t]}$}, i.e.,  	\begin{equation}
		\label{eq:aa2}
		\begin{split}
			P(\imgchanya|\predchanya,\ctxchanya)&=P(\imgchanya|\predchanya)\\
			\iff P(\imgchanya,\ctxchanya|\predchanya)&=P(\imgchanya|\predchanya)P((\ctxchanya|\predchanya))
		\end{split}
	\end{equation}
	
	\paragraph{Theorem 2: The Separable  Property of \TlI{} and \TvI{} \newline }
	\textit{Given assumption A2 holds, the effect of \tvi{} over label $P(\predchanya|\imgchanya)$ and the effect of $P(\predchanya|\ctxat{t})$   can be separated from contextual prediction  $P(\predchanya|\imgchanya,\ctxchanya)$}:
	\begin{equation}
		\begin{split}
			&P(\predchanya|\imgchanya,\ctxchanya)\\
			\propto&\frac{P(\predchanya|\imgchanya)P(\predchanya|\ctxchanya)}{P(\predchanya)}\\
			:=&Pr(\predchanya,\imgchanya,\ctxchanya)
		\end{split}
	\end{equation}
	
	\begin{proof} \label{prof:2}
		\begin{equation}
			\begin{split}
				&P(\predchanya|\imgchanya,\ctxchanya)\\
				=&\frac{P(\imgchanya,\predchanya,\ctxchanya)}{P(\imgchanya,\ctxchanya)}\\
				=&\frac{P(\imgchanya,\ctxchanya|\predchanya)P(\predchanya)}{P(\imgchanya,\ctxchanya)}\\
				\labelrel={prof:2:1}&\frac{P(\imgchanya|\predchanya)P(\ctxchanya|\predchanya)P(\predchanya)}{P(\imgchanya,\ctxchanya)}\\
				=&\frac{P(\predchanya)}{P(\imgchanya,\ctxchanya)}P(\imgchanya|\predchanya)P(\ctxat{t}|\predchanya)\\
				=&\frac{P(\predchanya)}{P(\imgchanya,\ctxchanya)}\
				\frac{P(\predchanya|\imgchanya)P(\imgchanya)}{P(\predchanya)}\
				\frac{P(\predchanya|\ctxchanya)P(\ctxchanya)}{P(\predchanya)}\\
				=&P(\predchanya|\ctxchanya)P(\predchanya|\imgchanya)\frac{P(\imgchanya)P(\ctxat{t})}{P(\imgchanya,\ctxat{t})P(\predchanya)}\\
				=&\frac{P(\predchanya|\imgchanya)P(\predchanya|\ctxchanya)}{P(\predchanya)}\frac{P(\imgchanya)}{P(\imgchanya|\ctxchanya)}\\
				=&Pr(\predchanya,\imgchanya,\ctxchanya)\frac{P(\imgchanya)}{P(\imgchanya|\ctxchanya)}\\
				=&Pr(\predchanya,\imgchanya,\ctxchanya)\frac{P(\imgchanya)}{\sum_{\predalterat{t}}^{\charset}P(\imgchanya|\predalterat{t},\ctxchanya)P({\predalterat{t}|\ctxchanya})}\\
				\labelrel={prof:2:2}&Pr(\predchanya,\imgchanya,\ctxchanya)\frac{P(\imgchanya)}{\sum_{\predalterat{t}}^{\charset}P(\imgchanya|\predalterat{t})P({\predalterat{t}|\ctxchanya})}\\
				\labelrel={prof:2:3}&\frac{Pr(\predchanya,\imgchanya,\ctxchanya)}{\sum_{\predalterat{t}}^{\charset} Pr(\predalterat{t},\imgchanya,\ctxchanya)}\\
				\labelrel\propto{prof:2:4}&Pr(\predchanya,\imgchanya,\ctxchanya).\\
			\end{split}
		\end{equation}
	\end{proof}
	Here, $\charset$ is the character set. Step \eqref{prof:2:1} and \eqref{prof:2:2} is derived using Eq. \ref{eq:aa2}  in assumption A2. Step~\eqref{prof:2:3} is derived by applying Bayesian rule over $P(\imgat{t}|\predalterat{t})$ and canceling $\imgat{t}$ . Although $\sum_{\predalterat{t}}^{\charset} Pr(\predalterat{t},\imgchanya,\ctxchanya)$ is not a constant number and may vary with timestamp $t$, but it is the same for all label $\predalterat{t}$ at a certain timestamp $t$, hence step \eqref{prof:2:4} holds, despite the constant factor can change. 
	\section{Proof of Theorem 3}
	\label{prof:3}
	Combining Theorem 1 and Theorem 2, we have the \Ca{} mechanism,
	\begin{equation}\label{thr}
		\begin{split}
			&P(\anypredseq|\imgword,l)\\
			=&\dcaright.\\
		\end{split}
	\end{equation}
	
	\begin{proof}
		\begin{equation}
			\begin{split}
				&P(\anypredseq|\imgword,l)=\prod_{t=1}^l \anchoreddist{P(\anypredat{t}|\imgat{t},\ctxat{t})P(\ctxat{t}|\imgword,l)}\\
				\propto&\prod_{t=1}^l \anchoreddist{\frac{P(\predchanya|\imgchanya)P(\predchanya|\ctxchanya)}{P(\predchanya)}P(\ctxat{t}|\imgword,l)}\\
				=&\alpha\prod_{t=1}^l\frac{P(\anypredat{t}|\imgat{t})}{P(\anypredat{t})}\prod_{t=1}^l\anchoreddist{P(\predchanya|\ctxchanya) P(\ctxat{t}|\imgword,l)},\\
				=&\dcarightfull,\\
				:=&\dcaright,\\
			\end{split}
		\end{equation}
	\end{proof} 
	The visual prediction can be taken out of the integral as it is not affected by the \tli{}, i.e., $\ctxat{t}$. Here, 
	\begin{equation}
		\beta(\mathbf{y})=\frac{\alpha}{\prod_{1}^{l}P(y_{t})},
	\end{equation}
	and it is a character frequency term related to the word. During training, $\beta(\mathbf{y^*})$ only associates with the label, hence would be constant for a certain label and won't produce gradient. 
	During the evaluation, as the dictionary and character frequency are unknown, character frequency would be assumed as uniform, resulting in $\beta(\mathbf{y})$ being a constant number $\frac{\alpha}{|C_{eval}|^l}$ for all words with length $l$.  Hence, despite varying from word to word, treating it as a constant does not affect either training or evaluation. As a result, $\beta$ is omitted for writing convenience.

\section{An Engineering Perspective of \JTRN}
While the main paper puts more stress on the theoretical part of the framework, we present an engineering perspective of the \JTRN{} framework which focuses on how things are implemented over what each module does.

\begin{figure}[!t]
	\centering
	\includegraphics[width=0.6\linewidth]{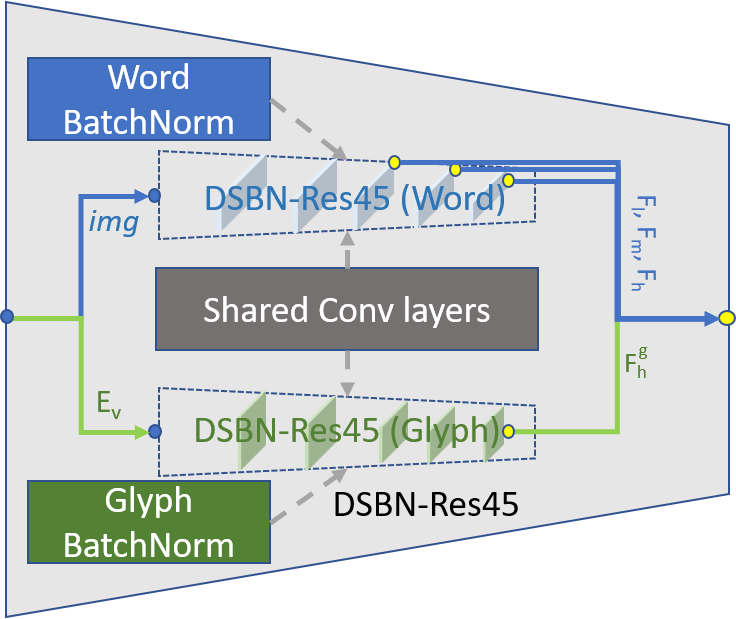}
	\caption{The DSBN-Res45 backbone.}
	\label{fig:res45}
\end{figure}

\subsection{DSBN-Res45}
In this work, the DSBN-Res45 (Fig.~\ref{fig:res45}) is used to encode word images and glyphs into corresponding visual features.   The DSBN-Res45 backbone is a modified version of the 45-layer ResNet used in \cite{DAN}. Here, we replaced all its batch norm layers with re-implemented DSBN~\cite{DSBN} layers. This adaption is made to alleviate the impact of the bias between the word image domain and the glyph domain. Specifically, the network uses the same set of convolution kernels for both word-level images and glyph images,  while using the domain-specific batch statistics for normalization of each specific domain. The layout of the regular model is similar to the original DAN implementation~\cite{DAN}, and the large model simply adds more latent channels to the backbone, and further details like specific network layout can be found in the released code. This module is a part of the base model and is used in \textbf{all} models in ablative studies.

\subsection{Prototype Generation}

\begin{figure}[!t]
	\centering
	\includegraphics[width=\linewidth]{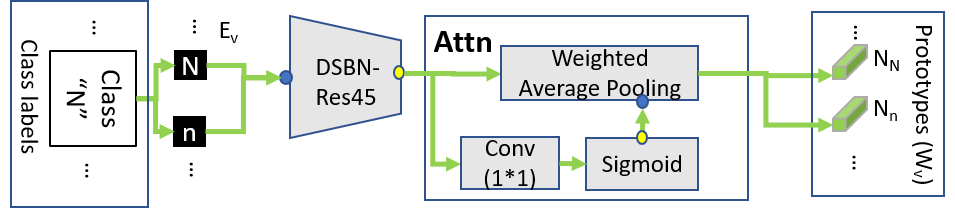}
	\caption{The prototype generation process.}
	\label{fig:proto}
\end{figure}

The prototypes generation process is shown in Fig.~\ref{fig:proto} for each class, the framework first extracts corresponding visual features of each glyph with the backbone. Then spatial attention is applied to reduce the feature map to a single feature via the Attn Module. Specifically, the module first estimates the foreground/background attention mask with a convolution layer, then reduces and normalizes the feature map into corresponding prototypes.  Same to ~\cite{neko20nocr}, a label may possess more than one prototype as each character can have different ``cases'', e.g., `N' and `n'.The prototypes are normalized to alleviate character-frequency related bias. 
This module is also a part of the base model and is used in \textbf{all} models in ablative studies. 

\begin{table*}[t]
	\caption{ Important notations in the paper with first occurrence and their brief explanations.}
	\begin{center}
		\begin{tabular}{c|c|p{13cm}}
			\hline
			Notation & Occurrence& Explanation\\
			\hline
			$N$ &-& Number of glyphs  (`N' and `n' has the same label)\\ 
			\hline
			$M$ &-& Number of characters (labels)\\   
			\hline 
			$E$ & Fig.2 & The collection of the glyph ($E_v$) and semantic embedding ($E_c$)\\
			\hline
			$E_c$ & Fig.2  & The collection of the glyph ($E_v$) for characters. Each character can have several glyphs according to how many cases it has.\\
			\hline
			$E_v$ & Fig.2  & The semantic embedding ($E_c$) for characters. Each character only has one embedding in our framework.\\
			\hline
			$W_v$ & Fig.2  & Prototypes generated from $E_v$. $W_v: R^{N\times D}$, $N=|E_v|$ and $D= 512$ \\
			\hline
			\hline
			$\predseq$ & Eq. 1 & Predicted word label, consisted of a ordered sequence of predicted character labels $\predat{t}$.\\
			\hline
			$\anyseq$ & Eq. 1 & Any word label, consisted of a ordered sequence of character labels $\anypredat{t}$.\\
			\hline
			$\predat{t}$ & Eq. 1 &$t^{th}$ predicted character.\\
			\hline
			$\theta$ &Eq. 1& The trainable parameters of the framework. \\
			\hline
			$\imgword$ &Eq. 1& \tvi{} of all characters in a word\\
			\hline
			$l$ &Eq. 2& Length of a sequence. \\
			\hline
			$\imgat{t}$ &Eq. 3& \tvi{} of the $t^{th}$ character in the word\\
			\hline
			$c$ &Eq. 3& Linguistic information. \\
			\hline
			$\gtseq$ &Eq. 4& Ground truth word label,  consisted of a ordered sequence of ground truth character labels $\gtat{t}$. \\
			\hline
			$l^{*}$ &Eq. 4& Length of the ground truth word. \\
			\hline
			$C_{[t]}$ &Eq. 4& All possible Linguistic information. \\
			\hline
			$L_{len}$ &Eq. 4& Minus likelihood of the correct length being predicted: $-logP(l^{*}|\imgword)$ \\
			\hline
			$L_{vis}$ &Eq. 4& Minus likelihood of the correct word being predicted according to \tvi{}: $-\sum_{t=1}^{l^{*}}(logP(\gtat{t}|\imgat{t}))$ \\
			\hline
			$L_{vis}$ &Eq. 4& Minus likelihood of the correct word being predicted according to \tli{}: 
			$-\sum_{t=1}^{l}log(\anchoreddist{P(\gtat{t}|\ctxat{t})P(\ctxat{t}|\imgword,l^{*}))}$ \\
			\hline
			$\anypredat{t}$ & Eq. 5 &Any $t^{th}$ character, applies to any possible character in the character set (means it applies to predicted and ground truth as well.)\\
			\hline
			\hline
			$\psi$ &Ch3.3.3& Function indexes all corresponding prototypes of the input label $\anypredat{t}$.\\
			\hline
			$w_v$ &Ch3.3.3& A row in $W_v$.\\
			\hline
			$\hat{c}$ &Ch3.3.4&Context  predicted  via transformer. $\hat{c}\in{R^{l\times D}}$\\
			\hline
			$\anydistat{t}$ &Ch3.3.4& The probability distribution at time stamp $t$: $P(\anydistat{t}|\imgat{t}): (P(\anypredat{t}^0|\imgat{t}),...,P(\anypredat{t}^M|\imgat{t}))$\\
			\hline
			$Y$ &Ch3.3.4& The probability distribution at all time stamps: \newline $Y\in (0,1)^{l\times M}:(P(\anydistat{0}|\imgat{0}),...,P(\anydistat{l}|\imgat{l}))$\\
			\hline
		\end{tabular}
		\label{tab:notation}
	\end{center}
\end{table*}
\subsection{Data, Training, and Evaluation} 
The training and evaluation and models for most experiments (all except dictionary-based close-set experiments) are now released to Kaggle\footnote{https://www.kaggle.com/vsdf2898kaggle/osocrtraining}.
For the open-set task, the training dataset is built by aggregating the following datasets: RCTW~\cite{rctw},  Chinese and Latin subset of the MLT-2019 dataset~\cite{mlt19}, LSVT~\cite{lsvt}, ART~\cite{art}, and CTW~\cite{ctw}. The training character set contains 3755 Tire-1 Simplified Chinese characters, 52 Latin characters, and 10 digits. Vertical Samples and samples containing characters outside of the training character set are removed from the training set (samples with Tradition Chinese Characters are removed as well). The evaluation dataset contains 4009 horizontal images from the Japanese subset of the MLT-2019 dataset and the testing character set includes $1460$ characters appearing in the evaluation set, making a total of $1461$ different classes adding the ``unknown'' class.  
For the close-set model, we use exactly the same datasets as DAN~\cite{DAN}, which adopts the most used MJ~\cite{MJ}-ST~\cite{ST} combo as  the  training set.
For the zero-shot Character recognition tasks, we reuse the split from OSOCR~\cite{neko20nocr}, which follows HCCR's protocol~\cite{hde}. Note that like HCCR~\cite{hde}, few methods made the exact split of seen and novel characters public.

During training, a label sampler is used to sample a subset of characters during each iteration like OSOCR. For word-level tasks, the model processes data is similar to DAN~\cite{DAN}. Specifically, the model takes 32*128 RGB clips, where the images are resized keeping aspect ratio and center-padded into 32*128 with zeros for both training and testing. The common dictionary-free protocols are used for all word-level evaluations. For character recognition tasks, the model treats character images like word images, despite the clip size being set as 32*64 to speed up. For all three tasks, we use Notofont as the glyph provider, where each character is rendered and centered to 32*32 binary patches.  
The training processes are mostly the same with DAN~\cite{DAN} except for adding a prototype generation process.

For evaluation, 
the most popular evaluation protocol, the Line Accuracy is used to measure word recognition performance following the community. The Character Accuracy (1-NED) is also used as compensation for open-set word recognition tasks to give an intuitive insight on the recognition quality per character. For the zero-shot Chinese Character recognition task, Character Accuracy and Line Accuracy is the same number, simply called Accuracy by the community. 

\section{Extra Details}
\subsection{Notations}
We made a notation table (Table \ref{tab:notation}) to include all used notations in this paper. 
In most cases, hat $(\hat{.})$ indicate the max probable prediction, and asterisk $(.^{*})$ indicates the ground truth. \textbf{Bold} notations indicate vectors and capital alphabets indicate matrices, sets, or distributions. 
\subsection{About Related Work}
\chgrevA{
	Despite the proposed \Ca{} mechanism (DCA) uses a transformer to  model contextual information, this work is not directly related to the transformer-based methods~ \cite{trans1,trans2,trans3}.
	Structure-wise, the transformer in our method is a BERT-style transformer encoder, also used in ~\cite{paddle,dsm}, while~\cite{trans1,trans2,trans3} uses GPT-style transformer decoders.}
\chgrevD{
	Purpose-wise, the transformer in this work is used as a regularization term (which is the direct reason we backpropagate to the feature encoder). Instead, ~\cite{paddle,dsm} use the transformer as a post-process module by cutting the gradient flow (Note ~\cite{paddle} did not clarify this in their paper, but you can see it from their officially released code). } \chgrevA{Conventional transformer-based methods directly decode the CNN features into prediction sequences.
	Therefore, the DCA module is different  in terms of structure and motivation.
}

\end{document}